\def\year{2020}\relax
\documentclass[letterpaper]{article} % DO NOT CHANGE THIS
\pdfoutput=1 % For ArXiV. This must appear in the first 2-3 lines.
\newcommand{\Buildkind}{arxiv}
\usepackage{lang-base}

\ifdefstring{\Buildkind}{blind}{
\pdfinfo{
/Title (Language-Level Automated Failure Analysis and Recovery For Service Mobile Robots)
/Author (Paper ID: 9379)
}
}
{
\pdfinfo{
/Title (Language-Level Automated Failure Analysis and Recovery For Service Mobile Robots)
/Author (Jenna Claire Hammond, Joydeep Biswas, and Arjun Guha)
}
}

\title{Automatic Failure Recovery for End-User Programs on Service Mobile Robots}

\ifdefstring{\Buildkind}{blind}{
  \author{Paper ID: 9379}
}{
\author{Jenna Claire Hammond\textsuperscript{\rm 1},
  Joydeep Biswas\textsuperscript{\rm 2}, and
  Arjun Guha\textsuperscript{\rm 3} \\
  Mount Holyoke College\textsuperscript{\rm 1},
  University of Texas at Austin\textsuperscript{\rm 2}, and
  University of Massachusetts Amherst\textsuperscript{\rm 3}
}
}

\begin{document}

\maketitle

\begin{abstract}

For service mobile robots to be most effective, it must be possible for
non-experts and even end-users to program them to do new tasks. Regardless of
the programming method (e.g., by demonstration or traditional programming),
robot task programs are challenging to write, because they rely on multiple
actions to succeed, including human-robot interactions. Unfortunately,
interactions are prone to fail, because a human may perform the wrong action
(e.g., if the robot's request is not clear). Moreover, when the robot cannot
directly observe the human action, it may not detect the failure until several steps
after it occurs. Therefore, writing fault-tolerant robot tasks is beyond the
ability of non-experts.

This paper presents a principled approach to detect and recover from a broad
class of failures that occur in end-user programs on service mobile robots. We
present a two-tiered \emph{\rtplfull{}} (\rtpl): 1)~an expert roboticist uses a
specification language to write a probabilistic model of the robot's actions
and interactions, and 2)~a non-expert then writes an ordinary sequential
program for a particular task. The \rtpl{} runtime system executes the task
program sequentially, while using the probabilistic model to build a Bayesian
network that tracks possible, unobserved failures. If an error is observed,
\rtpl{} uses Bayesian inference to find the likely root cause of the error, and
then attempts to re-execute a portion of the program for recovery.

Our empirical results show that \rtpl{} 1)~allows complex tasks
to be written concisely, 2)~correctly identifies
the root cause of failure, and 3)~allows multiple tasks to recover from
a variety of errors, without task-specific error-recovery code.

\end{abstract}

% =============================================================================
\section{Introduction}

The rapidly growing availability of service mobile robots has spurred
considerable interest in \emph{end-user programming} (EUP) for such
robots. The goal of EUP is to empower non-technical end-users to
program their service mobile robots to perform novel tasks. There are
now several ways to do so, including visual programming
languages~\cite{huang2016design,weintrop2017blockly},
natural language
speech commands~\cite{brenner2007mediating,roy2000spoken,duvallet2013imitation},
and other domain-specific languages~\cite{mericcli2014interactive}. However,
irrespective of the
the mode of entry (be it speech, visual, or textual DSLs), EUP
remains hard because service mobile robots inevitably encounter errors,
and writing \emph{robust} programs that can cope with failures is a
challenging problem.

Robot task programs often consist of sequences of actions where later actions
depend on the successful execution of former actions, but such dependencies are
rarely specified formally. Furthermore, the outcomes of certain kinds of
actions\,--\,especially actions where the robot asks for human
assistance\,--\,may not be immediately observable by the robot. Therefore, it
may take several steps of execution before the robot can detect
that a past action failed. For a non-expert, who is not trained to reason
about a robot's failure modes, and may have limited programming
experience, writing robust programs is challenging because the number of
possible combinations of recovery steps quickly grows large.

This paper addresses these challenges with end-user programming for service
mobile robots. We present a new domain-specific language,
\rtplfull{} (\rtpl{}) that provides end-users with a familiar syntax for writing
robot tasks. During execution, the \rtpl{} runtime system builds an internal
representation that
\begin{inparaenum}[1)]
  \item explicitly tracks the dependencies of every action in the task program;
  \item reasons about the outcome of every action probabilistically;
  \item when it encounters errors, runs probabilistic inference to find the most likely cause of the errors; and
  \item when possible, autonomously executes recovery steps to transparently overcome such errors.
\end{inparaenum}
Our empirical results show that \rtpl{} 
\begin{inparaenum}[1)]
\item automatically recovers from a wide range of errors without the need for
explicit failure inference and recovery;
\item infers the most likely causes of failures depending on the probabilistic
model of the robot's actions; and
\item recovers from errors encountered by a real service mobile robot,
allowing it to complete tasks faster than na\"ive re-execution.
\end{inparaenum}

% =============================================================================
\section{Related Work}

There are several approaches to EUP for robots,
with input modalities ranging from programming by
demonstration~\cite{alexandrova2014robot}, natural
language~\cite{brenner2007mediating,roy2000spoken,duvallet2013imitation}, and
virtual reality interfaces~\cite{featherston2014dorothy}. Moreover, these
programs can be represented in a variety of ways, including
blocks~\cite{weintrop2018evaluating,huang2016design}, instruction
graphs~\cite{mericcli2014interactive}, and state abstractions~\cite{cobo2011automatic}. Irrespective of the
mode of program input and its representation, writing robust programs that can
handle failures remains a challenging problem. Our approach to automated
failure recovery is orthogonal to both input modality and program
representation, as long as the end-user programs are \emph{sequential} in
nature (\secref{rtpl})\,--\,that is, they do not consist of parallel execution paths.

%Learning from demonstration
Learning from demonstration (LfD) has been used extensively to teach
robots new low-level motor skills, such as manipulation~\cite{kroemer2019review},
navigation~\cite{ellis2013autonomous}, and
locomotion~\cite{gonzalez2013humanoid}. The representation of the LfD-learned policy may
vary, but it is most commonly 
not intended to be human-comprehensible. In contrast, we focus on novel task
programs represented in a form (\eg{} source code) that is comprehensible to the
end-user, to aid in modifications, re-use, or re-parameterization.

Rousillon~\cite{rousillon} uses programming by demonstration (PbD)
to synthesize web-scraping programs that are robust to
failures that may arise due to format and layout changes on websites. In
contrast, our work reasons probabilistically about failures and allows
automatic failure recovery, even when the failures are not directly observable
by the robot.

Automated task planning, while successful in domains with well-specified
problems~\cite{wray2017online,brechtel2011probabilistic}, is not well-suited
for end-user programming. Planning-based EUP would formal specifications of
task goals, which vary significantly and is known to be challenging even for
experts~\cite{actl-vacuity}. Therefore, our work focuses on working directly
with tasks that have already been implicitly specified in terms of the
necessary sequence of actions.

\emph{Plan repair}~\cite{van2005plan} is closely related to automated planning,
and allows a plan to be modified during
execution in light of new constraints, by using local refinements. Plan repair
again requires a formal specification of goal conditions, and a separate
inference algorithm for detecting failures. In contrast, \rtpl{} does
\emph{both} probabilistic inference of the most likely causes of failure
\emph{and} synthesizes repairs, without formal specifications for the task.
Aside from generic plan repair, there has been some work on repairing robot
behaviors, including state transition functions~\cite{holtz2018interactive} and
control systems~\cite{mericcli2012multi}. However, such approaches rely on either human
corrections, or hand-crafted recovery procedures. In contrast, our proposed
approach \emph{autonomously} generates repairs for end-user programs, without
the need for either human corrections or hand-crafted recovery procedures.

% =============================================================================
\section{The \rtplfull{}}
\seclabel{rtpl}

\begin{figure}[tb]
% \lstset{language=python}
\begin{lstlisting}
robot.goto("mail room")
robot.prompt("Please place the packages for A and B in my basket.")
robot.goto("location A")
robot.prompt("Please take the package for A.")
robot.goto("location B")
robot.prompt("Please take the package for B.")
\end{lstlisting}
  \caption{A canonical robot task program to pickup and deliver two packages.
  Every action can fail, including the actions that involve interacting with humans.}
  \figlabel{pickup-2-no-errors}
\end{figure}

\lstset{language=PDDL,style=normalstyle}
This section presents \rtplfull{} (\rtpl). As a running example, we consider a
service mobile robot that can autonomously navigate in an environment, and
interact with the human occupants. The robot is not equipped with arms to
manipulate objects, but can request help from humans to manipulate objects, for example
to place packages in its basket, or to pick them up.

\figref{pickup-2-no-errors} shows a canonical example of an end-user program
for such a robot: the robot goes to the mail room to pick up two packages
and then delivers them to two locations. In this program,  pickup and delivery
are human interaction actions, where the robot proactively prompts a human
for assistance. Unfortunately,
every line in this task program can go wrong. The \lstinline|goto| actions may
fail if the robot's path is entirely blocked; the humans in the mailroom may fail to
give the robot one or both packages; someone at location $A$ may pickup the
package for location $B$; if there is nobody at a location to pickup the
package, the robot will wait indefinitely; and so on. Furthermore, this program
conceals the implicit dependencies between actions, e.g., the robot
can only deliver the package to location $A$ if the human successfully gives the
package to the robot in the mail room. Therefore, a robust
implementation of this task must be significantly more complicated to address
these and other contingencies.

The goal of \rtpl{} is to allow non-experts to write task programs that are
as straightforward as the one in \figref{pickup-2-no-errors}, but can exhibit complex
behaviors to recover from errors. \rtpl{} is a \emph{two-tiered programming
language}, thus a complete program consists of two parts that serve different
roles:

\begin{enumerate}

\item \emph{Robot Model:} The expert roboticist writes the first-tier
program, which is a declarative specification of the actions that the robot can
perform, their nominal behavior including parameterized preconditions and effects, and a
probabilistic model of possible failures. The parameters to each action
(\eg, \lstinline|PackageA| as the parameter to
\lstinline[mathescape]|robot.give($\cdot$)|), along with the action
preconditions and effects, formally encode the dependencies in a task program.
The robot model enables probabilistic inference of the most likely cause of
failures when a failure occurs in a task program.

\item \emph{Task Program:} A non-expert writes the second-tier program, which
is an ordinary sequential program that makes the robot perform some task. This
program uses the actions specified in the first-tier, thus benefits from
automatic failure detection and recovery (\secref{failure-recovery}). We
present programs written in Python, but our approach will work with any
sequential language, including non-textual
languages.

\end{enumerate}

A feature of this design is that the a single expert-written robot model can
endow a variety of robot task programs with automatic failure recovery, without
the need for any task-specific failure recovery code.
Moreover, it is possible for the robot model and the task program to evolve
independently. Over time, the expert may update the robot model with better
priors or even new kinds of failures, without requiring task programs to
change. In the rest of this section, we first present how experts write robot
models and then show how non-experts write task programs.

\begin{figure}
\begin{subfigure}{\columnwidth}
\begin{lstlisting}[language=pddl]
(:action enter-room
  :parameters (?r - room)
  :precondition
    (and (door-open (door r))
         (at (outside r)))
  :postcondition
    (and (at (inside r))
         (not (at (outside r))))
  :belief-update enter_room_bupdate)
\end{lstlisting}
\caption{Nominal specification.}
\figlabel{enterroom_pddl}
\end{subfigure}

\begin{subfigure}{\columnwidth}
\begin{lstlisting}[language=python]
def enter_room_bupdate(w, r):
    in_loc = inside(r)
    out_loc = outside(r)
    w_next = w.clone()
    w_next.at[in_loc] = $(1 - \alpha) $w.at[out_loc]
    w_next.at[out_loc] = $\alpha$ w.at[out_loc]
    return w_next
\end{lstlisting}
\caption{Probabilistic specification. $\alpha$ is the probability that the door is detected open incorrectly.}
\figlabel{enterroom_ppl}
\end{subfigure}

\caption{\rtpl{} specification of the \texttt{enter-room} action.}
\end{figure}

\subsection{Expert-Provided Robot Model}
\seclabel{robot-model}
\lstset{language=pddl}

The first tier of \rtpl{} is a domain-specific language (DSL) for specifying
the actions that the robot can perform. We use an extension of Planning Domain
Definition Language (PDDL)~\cite{pddl_original} to specify actions. Every
action has a name, a list parameters, a precondition, a postcondition, and
a belief update function.

The \emph{parameters} of an action all have a name and a type, e.g.
\lstinline|location|, \lstinline|room|, or \lstinline|door|. The set of types
is straightforward to extend and the execution and failure recovery algorithms
work with arbitrary type definitions.

Every action has a \emph{precondition} and \emph{postcondition} that hold
under nominal execution. 
To make task programs easier to write, we allow the robot model to use simple
functions that map from one type to another.\footnote{This feature is an
extension to PDDL, which requires function-free first-order logic. However,
they do not affect our approach to failure recovery, because it only searches
previously executed actions, not the space of all possible actions.} For
example, the \lstinline|(inside ?r)| function returns a fixed location inside
the room \lstinline|?r| and the \lstinline|(door ?r)| function returns a ID of
the door to room \lstinline|?r|.

Finally, every action names a \emph{belief update function}, which is defined separately
as an ordinary function in code. The belief update function takes as its
arguments 1)~a probabilistic world state \lstinline|w| and 2)~the action parameters, and
returns a distribution of worlds. The probabilistic world state (explained in
depth in \secref{execution}) assigns a probability to the likelihood of each
literal in the world state being true. For example, \lstinline|w.at[$x$] == 0.9| holds if
the robot is at location $x$ with probability 0.9. Any literal that is not defined in
the world state is assumed to be identically false.
The \rtpl{} runtime system uses the belief update
function to build a probabilistic model of world state.

Note that the pre- and post- conditions do not account for real-world execution errors. For example,
door detection is imperfect: a temporary obstacle, such as a person walking
past an open door, can fool the sensor into incorrectly reporting that the door
is closed. For example, the \lstinline|enter-room| action (\figref{enterroom_ppl})
has a belief update function that uses the parameter $\alpha$, which determines
the prior probability that the door
detector incorrectly reports that the door is open.

\begin{figure}
\begin{subfigure}{\columnwidth}
\begin{lstlisting}[language=pddl]
(:action pickup
  :parameters (?l - location ?x - item)
  :precondition (at ?l)
  :postcondition (have ?x)
  :belief-update pickup_bupdate)
\end{lstlisting}
\caption{Nominal specification.}
\figlabel{pickup_pddl}
\end{subfigure}

\begin{subfigure}{\columnwidth}
\begin{lstlisting}[language=python]
def pickup_bupdate(w, l, x):
  w_next = w.clone()
  w_next.have[x] = $1 - \alpha$
  return w_next
\end{lstlisting}
\caption{Probabilistic specification. $\alpha$ is the probability of the human accidentally failing to give the item to the robot.}
\figlabel{pickup_ppl}
\end{subfigure}
  
\caption{\rtpl{} specification of the \texttt{pickup} action.}
\end{figure}

\paragraph{Human-Robot Interaction Specifications}

A key feature of \rtpl{} is that it can describe
human-robot interactions using the same DSL that we use to describe autonomous
actions.
\figref{pickup_pddl} shows the nominal specification of the \lstinline|pickup|
interaction, which directs the robot to ask a human to give it an item
(\lstinline|?x|) at a location (\lstinline|?l|). The precondition requires the
robot to be at the location and the postcondition states that the robot has the
item if the action succeeds. To model possible errors, we complement the nominal
specification with a probabilistic specification (\figref{pickup_ppl}). In this
specification, the parameter $\alpha$ is the prior probability that the robot
does not have the item even if the human confirms that it has given the robot
the item. 

In practice, service mobile robots can only perform a limited set of actions
autonomously. However, there is a much broader variety of human-robot
interactions that make service mobile robots far more versatile than just their
autonomous capabilities allow~\cite{rosenthal:symbiotic}. We have used
\rtpl{} to specify nine typical human-robot interactions for the service
mobile robot that we have in our lab, and used them to build a variety of
task programs which we present in \secref{evaluation}.
 These actions are straightforward to specify
and follow the same pattern employed by \lstinline|pickup|: each action has
an independent parameter that determines the probability that the human
action succeeds or fails.

% Figures for Section 2.2 below
\begin{figure}
\lstset{language=python}
\begin{lstlisting}
robot.goto("mail room")
robot.pickup("Package A")
robot.pickup(Package B")
robot.goto("location A")
robot.give("Package A")
robot.goto("location B")
robot.give("Package B")
\end{lstlisting}
\caption{A 2-package delivery program written in \rtpl{}.}
\figlabel{pickup-2-automatic}
\end{figure}

\subsection{Task Programs}
\seclabel{task-programs}
\lstset{language=python}

The defining characteristic of an \rtpl{} task program is how it performs human
interactions. An \rtpl{} program invokes an action, such as
\lstinline|pickup|, and the implementation of the action
abstracts the low-level system code needed to display a human-readable prompt
along with buttons that allow the human to confirm that they completed the
action or indicate that they cannot do so.\footnote{\rtpl{} does not require all interactions to use the robot
model. However, an ad hoc interaction does not get recorded by the
\rtpl{} runtime system, thus does benefit from automatic failure
recovery.}  \rtpl{} starts failure recovery
when a human indicates that they cannot perform an action (\secref{failure-recovery}).

Unlike in previous approaches for programming service mobile robots where interaction
strings are only used for displaying on screen or for speech synthesis, \rtpl{}
\textbf{simultaneously uses such parameters to explicitly encode dependencies of
human interaction actions}. Moreover, \rtpl{} supports dynamic implicit arguments: if an
action in the task program omits an argument (\eg{} the location of the
\lstinline|pickup()| action), then its value is inferred from
the current world state (the mail room). Thus, where in previous designs there was no way to automatically reason
about the dependencies of
actions that rely on human-performed
actions, the \rtpl{} approach enables task programs to \emph{automatically}
recover from failures. This is possible because the action specifications in the robot model
include a formal definition of the pre- and post-conditions in terms of the
parameters of the actions, along with a probabilistic model of likely errors in
terms of the specified parameters. \figref{pickup-2-automatic} shows the
equivalent program of \figref{pickup-2-no-errors}, written in \rtpl{}. The
parameters to the actions are used by \rtpl{} to automatically infer
dependencies, for example that the \lstinline|robot.pickup("Package A")| will
result in the robot satisfying the precondition that the robot have the package
for the later action \lstinline|robot.give("Package A")|.

% =============================================================================
\section{Nominal Execution and Failure Detection}
\seclabel{execution}

This section describes how \rtpl{} operates during normal execution, which
includes failure detection. The next section presents failure recovery. The
\rtpl{} runtime system maintains an explicit estimate of the robot's state. A
conventional STRIPS-style representation of the world state $W$ would consist
of a conjunction of $n$ propositional literals, $W = \wedge_{i=1}^{i=n} x_i$,
where any literal not included in the world state is assumed to be false.

However, a STRIPS representation is deterministic thus it cannot
capture the probabilistic nature of actions. Therefore, we introduce
the \emph{Bernoulli-STRIPS State Representation} (\bssr), which tracks robot
state and accounts for probabilistic effects. In \bssr{}, every literal
is associated with a  corresponding random variable $p(x_i)$ drawn from a
Bernoulli distribution, such that $p(x_i)=y_i$ implies that the literal $x_i$ is
true with probability $y_i$, and false with probability $(1-y_i)$. Since each
literal in the world state tracks distinct event outcomes, we assume their
corresponding Bernoulli random variables are independent:
$p(x_i,x_j)=p(x_i)p(x_j) \forall i\neq j$.
Thus, the probabilistic world state $p(W)$ in a \bssr{} is given by,
\begin{align}
  p(W) = \prod_{i=1}^{i=n} p(x_i).
\end{align}
As in STRIPS, a literal that does not exist in the world state in \bssr{} is assumed to be
false: $x' \notin W \Rightarrow p(x')=0$. 

\paragraph{Maximum Likelihood and Predicate Evaluation}

Given a \bssr{} literal $p(x)$, the \emph{maximum likelihood} value of the
corresponding STRIPS literal $x^* = \ML[p(x)]$ is $\true$ iff $p(x)>0.5$, and
$\false$ otherwise -- this follows from each \bssr{} literal being drawn from a
Bernoulli distribution. The maximum likelihood operator $\ML[\cdot]$ is similarly
defined to evaluate the maximum likelihood world state $W^*$ from a \bssr{}
world state $p(W)$ as
\begin{align}
  W^* &= \ML[p(W)] \nonumber \\
      &= \wedge_{i=1}^{i=n}\ML[p(x_i)].  
\end{align}
During execution, \rtpl{} needs to evaluate preconditions against the
current \bssr{} world state. Given a predicate $r=\wedge_{j=1}^{j=m}x_j$ for the
precondition of the next action, \rtpl{}
evaluates its maximum likelihood $r^*$ from the maximum likelihood value of the corresponding \bssr{} variables in
the world state, accounting for whether each literal is present in the world
state or not:
\begin{align}
  r^*&=\wedge_{j=1}^{j=m}x_j^*, \nonumber \\
  x_j^*&= \left\{ 
    \begin{array}{ll}
      \ML[p(x_j)] & \text{if } x_j \in W \\
      \false & \text{else}
    \end{array}
    \right.
  \eqlabel{bssr-predicate}
\end{align}
If the maximum likelihood of the predicate $r^*$ evaluates
to true given a \bssr{} world state $p(W)$, then we denote the implication as
$p(W) \Rightarrow r$.

\paragraph{Action Execution}

The initial world state ($W^0$) at the start of the program consists of an empty
set of literals, just as in STRIPS plan execution, $W^0=\varnothing$. The
corresponding \bssr{} world state is identically true: $p(W^0)=1$. To execute a
single action with precondition $r$ and belief-update function $f$ in the
\bssr{} world state $p(W^t)$, \rtpl{} proceeds in three steps:

\begin{enumerate}

\item It evaluates the precondition in the current world state ($r|_{p(W^t)}$).

\item If the precondition is
true, it performs the action (which may be a human interaction).

\item If the action succeeds, it uses the belief-update function to calculate
the updated \bssr{} state ($p(W^{t+1}) = f(W^t)$). The updated \bssr{} state
will have updated values for the Bernoulli distributions of a subset of its
predicates from $W^t$, as determined by the belief-update function $f$ of that action. Moreover,
it may even define new \bssr{} literals that were not present in $W^t$.

\end{enumerate}

\begin{figure}

\centering
\tikzset{
  x=1cm,
  y=-1cm,
  style={font=\footnotesize,node distance=1cm},
  arr/.style={-latex},
  world/.style={circle,draw, node distance=1.2cm},
  action/.style={circle,draw,node distance=0.5cm,font=\tiny},
  action_lab/.style={right,xshift=0.5cm,yshift=0.5cm},
  action_prob/.style={right},
  world_lab/.style={left,xshift=-0.3cm},
}

\begin{tikzpicture}[scale=0.8, every node/.style={scale=0.8}, every text node part/.style={align=center}]
\node[world] (w0) {$W^0$};
\node[world,below of=w0] (w1) {$W^1$};
\node[world,below of=w1] (w2) {$W^2$};
\node[world,below of=w2] (w3) {$W^3$};
\node[world,below of=w3] (w4) {$W^4$};
\node[world,below of=w4] (w5) {$W^5$};

\node[action_lab] (a1_lab) at (w0.east) {goto(M)};
\node[action_lab] (a2_lab) at (w1.east) {pickup(A)};
\node[action_lab] (a3_lab) at (w2.east) {pickup(B)};
\node[action_lab] (a4_lab) at (w3.east) {goto(A)};
\node[action_lab] (a5_lab) at (w4.east) {give(A)};

\node[action, below of=a1_lab] (a1) {$a^1$};
\node[action, below of=a2_lab] (a2) {$a^2$};
\node[action, below of=a3_lab] (a3) {$a^3$};
\node[action, below of=a4_lab] (a4) {$a^4$};
\node[action, below of=a5_lab] (a5) {$a^5$};

\node[action_prob] (a1_prob) at (a1.east) {$1-\alpha_1$};
\node[action_prob] (a2_prob) at (a2.east) {$1-\alpha_2$};
\node[action_prob] (a3_prob) at (a3.east) {$1-\alpha_2$};
\node[action_prob] (a4_prob) at (a4.east) {$1-\alpha_1$};
\node[action_prob] (a5_prob) at (a5.east) {$1-\alpha_3$};

\node[world_lab] at (w1.west) {$p(\mathrm{at}_M)$ \\ $1-\alpha_1$};
\node[world_lab] at (w2.west) {$p(\mathrm{at}_M),p(\mathrm{have}_A)$ \\
$1-\alpha_1 , 1-\alpha_2$};
\node[world_lab] at (w3.west)
{$p(\mathrm{at}_M),p(\mathrm{have}_A),p(\mathrm{have}_B)$ \\
$1-\alpha_1 , 1-\alpha_2 , 1-\alpha_2$};
\node[world_lab] at (w4.west)
{$p(\mathrm{at}_M),p(\mathrm{have}_A),p(\mathrm{have}_B),p(\mathrm{at}_A)$ \\
$\alpha_1-\alpha_1^2, 1-\alpha_2 , 1-\alpha_2, 1-(\alpha_1-\alpha_1^2)$};
\node[world_lab] at (w5.west)
{$p(\mathrm{at}_M),p(\mathrm{have}_A),p(\mathrm{have}_B),p(\mathrm{at}_A)$  \\
$\alpha_1-\alpha_1^2, \alpha_3-\alpha_3\alpha_2 , 1-\alpha_2, 1-(\alpha_1-\alpha_1^2)$};

\draw[arr] (w0) -- (w1);
\draw[arr] (w1) -- (w2);
\draw[arr] (w2) -- (w3);
\draw[arr] (w3) -- (w4);
\draw[arr] (w4) -- (w5);
\draw[arr] (a1) -- (w1);
\draw[arr] (a2) -- (w2);
\draw[arr] (w2) -- (a3);
\draw[arr] (a3) -- (w3);
\draw[arr] (a4) -- (w4);
\draw[arr] (w4) -- (a5);
\draw[arr] (a5) -- (w5);

\end{tikzpicture}
\caption{An example Bayes net constructed during normal execution of an \rtpl{}
task program.}
\figlabel{bayesnet}
\end{figure}
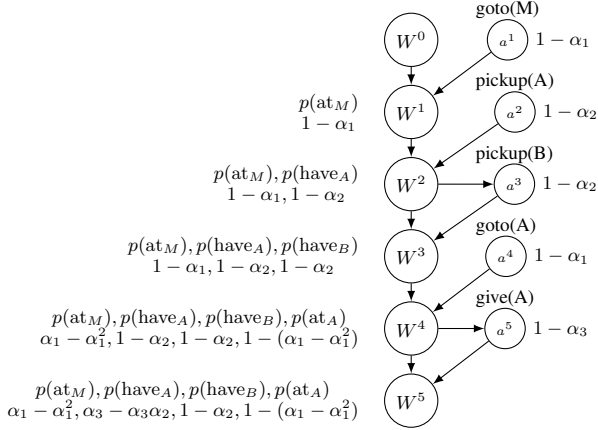

As the \rtpl{}
program executes, it incrementally builds a Bayes net of the \bssr{} predicates 
over time, depending on the sequence of actions performed. For each time-step $t_i$ of
the task program, the Bayes net includes variables for the \bssr{} world state
$p(W^i)$ from that time-step. Each action $a^i$ from timestep $t_i$ may include
additional action-specific variables, depending on the belief update function of
the specific action. For example, the action
\lstinline[mathescape]|robot.pickup($\cdot$)|, presented in \figref{pickup_ppl},
results in the addition of a variable to track whether the human gave the
package to the robot or not. In general, each action $a^i$ introduces
$m_i$ action-specific variables $a^i_j:j\in[1,m_i]$ to the Bayes net.
\figref{bayesnet} 
shows an example Bayes net constructed for the \rtpl{} program listed in 
\figref{pickup-2-automatic}. 

\paragraph{Failure Detection}

As mentioned above, there are two ways in which the execution of an action can
fail. First, if the precondition of an action is false in the current \bssr{}
world state, then \rtpl{} triggers error recovery~(\secref{failure-recovery})
before attempting to execute the action. This allows task programs to
seamlessly handle failures that arise independent of human interactions, such
as finding that a door to a room is closed before the robot tries to enter via
the door.

The second kind of failure occurs when the action's precondition holds, but the
action nevertheless fails. This can occur because of observation error or
because an action, especially a human interaction, may not be directly
observable by the robot. In these cases, the failure includes as evidence
values for literals in the precondition that make the precondition fail. These
values are used as observations for backward inference in recovery, which
we present in the next section.

% =============================================================================
\section{Backward Inference and Failure Recovery}
\seclabel{failure-recovery}

%\jb{\textbf{1.} Bayesian backward inference.}\\
During execution of a user program, \rtpl{} builds a time-indexed Bayes net
that relates the \bssr{} world states from all previous time-steps $p(W^{0:t})$,
along with action-specific variables $a_j^{1:t}$ for each time-step. When a
failure is detected, the evidence for the failure, which is a STRIPS
predicate $e_f$, is used to perform full a-posteriori inference over all
previous world states, conditioned on the evidence: $p(W^{0:t}|e_f)$. We use
standard variable elimination for this inference -- as shown in
\secref{execution}, the Bayes net from forward execution has a linear pattern
dictated by the program execution trace, which makes the inference particularly
conducive to computationally efficient inference via variable elimination.

% \jb{\textbf{2.} Find the first time-step where the maximum likelihood STRIPS
% world state changed.}\\
Given the inferred previous world states conditioned on the evidence
$p(W^{0:t}|e_f)$, to determine the first likely time-step that caused the
failure, \rtpl{} finds the first time-step $t_f$ where the maximum
likelihood world state conditioned on the failure evidence differs from the maximum
likelihood forward-predicted world states:
\begin{align}
  t_f = \argmin_i \ML[p(W^i|e_f)] \neq \ML[p(W^i)]
\end{align}
Thus, at time-step $t_f$ there will be one or more literals that differ between
the a-posteriori, and the forward-predicted world states, and thus comprise the
\emph{failure predicate set} $r_f$ given by,
\begin{align}
  r_f = \{x_j : x_j \in W^{t_f}, \ML[p(x_j|e_f)] \neq \ML[p(x_j)]\}.
\end{align}

%\jb{\textbf{3.} Check if the failure is recoverable.}\\
Depending on the failure predicate set $r_f$, the cause of the failure could be because of one of two possible
cases:
\begin{enumerate}
\item \textbf{Postcondition failure:} An expected postcondition of an action was
failed to be satisfied (\eg{} the human forgot to give the package to
the robot when asked), or
\item \textbf{Unintended effects:} An action resulted in an unintended
abnormal effect (\eg{} a human accidentally picked up the wrong package).
\end{enumerate}
Note that a \emph{precondition failure}, where an action's preconditions are
not met, must necessarily be preceded by either a postcondition failure or an
unintended effect, if the user program is a valid executable \rtpl{} program. 

Failures that result from postcondition failures may be automatically
recoverable by the \rtpl{} runtime, and they trigger the failure recovery
procedure presented in \secref{failure-recovery}.
Failures from unintended effects are not autonomously recoverable -- in fact,
they may not be recoverable at all (\eg{} if a package gets stolen from the
robot during transit). In such unrecoverable failures, the \rtpl{} runtime
reports the inferred cause of the failure to the user, and aborts execution.

% \subsection{Failure Recovery by Perforated Execution}
% \seclabel{failure-recovery}

In the case where a postcondition failure is estimated to be the cause of the
failure, the \rtpl{} runtime attempts failure recovery by re-executing the
action $a^{t_f}$ from that time-step. Unfortunately, the robot cannot directly
execute action $a^{t_f}$, since it may require preconditions that may no longer
be valid. Thus, the \rtpl{} runtime first determines which past
actions need to be re-executed in order to satisfy the preconditions for
$a^{t_f}$. The actions to be re-executed form a \emph{perforated trace} $\petrace$,
represented as a vector of binary indicator variables $b_i\in\{1,0\}$
$\petrace=\langle b_1,\ldots{},b_i,\ldots{}b_{t_f}\rangle$ that indicate whether the
corresponding actions $a^i$ from time-step $t^i$ should be executed or not.  A \emph{valid} perforated trace is one such that the
preconditions of every action $a^i$ in it must be satisfied by the world state
at that time-step: $W_f^i \Rightarrow a^i.\texttt{pre}$, where $W_f^i$ is the
world-state at time-step $i$ of the perforated execution.
The \emph{length} of a perforated trace ($||\petrace||$) is defined as
the number of actions that 
need to be re-executed by it: $||\petrace|| = \sum_{i=1}^{i=t_f}b_i$.
The goal of automated repair by the \rtpl{} runtime is thus to find a valid perforated
trace $\petrace$ of minimum length such that the final repaired world state
$W_f^{t_f}$ satisfies the preconditions of the final repair action $a^{t_f}$:
\begin{align}
  \petrace^* =& \argmin_{\petrace} ||\petrace|| \textrm{ s.t. } \\
   & \forall i\in[0,t_f-1], W^{i+1}_f = \left\{ \begin{array}{ll}
    a^i(W^i_f) &\text{if } b_i\\
    W^i_f &\text{else}
   \end{array}\right. \\
   & \forall i\in[0,t_f], W^i_f \Rightarrow a^{i}.\texttt{pre}.
\end{align}
Searching for the valid, optimal perforated trace $\petrace^*$
is closely related to the backward search step in
GraphPlan~\cite{blum1997fast,kambhampati2000planning}, but while GraphPlan may have several
(potentially mutexed) possible actions at every step, our search problem
only considers two options at each time-step $i$: either
executing $a^i$, or the persistence action (not executing $a^i$). Thus, the
search space for the optimal valid perforated trace is significantly smaller
than in general GraphPlan backward search.

% \jb{\textbf{4.} Present recovery algorithm: partial re-execution of the guarded execution trace built so far.}\\

% ==============================================================================
% Evaluation section
\section{Evaluation}
\label{sec:evaluation}

This section evaluates \rtpl{} by answering three questions. 1)~Can a single
task program demonstrate a variety of failure recovery behaviors
that would normally require complex logic? 2)~How do parameter values, which are
set by the expert roboticist, affect error recovery in a task program written by
a non-expert? 3)~Does \rtpl{} save time in practice?

\begin{table}[t]
\tikzset{
  x=1cm,
  y=-1cm,
  style={font=\footnotesize,node distance=1cm},
  ok_arr/.style={-latex},
  err_arr/.style={thick,-latex,color=red},
  err_node/.style={fill=red!50},
  unrec_node/.style={cross out,red,minimum width=0.5cm,draw},
  ok_node/.style={circle,minimum width=0.5cm,draw},
  label/.style={node distance=0.5cm,anchor=south,align=center},
  edge_label/.style={midway,below,font=\tiny},
}
 
\centering
\begin{tabular}{|c|l|}
\hline
\thead{EUP} &
\thead{Execution Trace}
\\ \hline

2-PD
&
\renewcommand{\arraystretch}{0.2}% Tighter
\begin{tabular}{@{}l}
\rule{0pt}{3.5em}
\begin{tikzpicture}[scale=0.8, every node/.style={scale=0.8}]
% Nodes that represent states
\coordinate (begin) {};
\node[ok_node,dashed,right of=begin,xshift=-0.25cm] (goto_office) {};
\node[ok_node,right of=goto_office] (pickup1) {};
\node[ok_node,right of=pickup1] (pickup2) {};
\node[ok_node,dashed,right of=pickup2] (goto1) {};
\node[ok_node,right of=goto1] (deliver1) {};
\node[ok_node,dashed,right of=deliver1] (goto2) {};
\node[ok_node,err_node,right of=goto2] (deliver2) {}; % Use ok_node,err_node for errors

% Nodes that are labels below states
\node[label,below of=goto_office] {$G$};
\node[label,below of=pickup1] {$P_A$};
\node[label,below of=pickup2] {$P_B$};
\node[label,below of=goto1] {$G$};
\node[label,below of=deliver1] {$D_A$};
\node[label,below of=goto2] {$G$};
\node[label,below of=deliver2] {$D_B$};

\coordinate[above of=deliver2,node distance=0.7cm] (anchor1);
\coordinate[above of=pickup2,node distance=0.7cm] (anchor2);
% All the arrows
\draw[ok_arr] (begin) -- (goto_office);
\draw[ok_arr] (goto_office) -- (pickup1) node[edge_label]{1};
\draw[ok_arr] (pickup1) -- (pickup2) node[edge_label]{2};
\draw[ok_arr] (pickup2) -- (goto1) node[edge_label]{3};
\draw[ok_arr] (goto1) -- (deliver1) node[edge_label]{4};
\draw[ok_arr] (deliver1) -- (goto2) node[edge_label]{5};
\draw[ok_arr] (goto2) -- (deliver2) node[edge_label]{6};
\draw[err_arr] (deliver2) -- (anchor1) -- (anchor2) node[xshift=0.5cm,edge_label]{7} -- (pickup2);
\draw[err_arr] (pickup2) edge[out=40,in=150] node[edge_label]{8} (goto2);
\draw[err_arr] (goto2) edge[out=30,in=150] node[edge_label,above]{9} (deliver2);
\end{tikzpicture} \\
\scriptsize Package B missing, robot picks it up again.
\end{tabular}

\\ \hline
%
% Distributor forgot to give package 1
2-PD
&
\renewcommand{\arraystretch}{0.2}% Tighter
\begin{tabular}{@{}l}
\rule{0pt}{3.0em}
\begin{tikzpicture}[scale=0.8, every node/.style={scale=0.8}]
% Nodes that represent states
\coordinate (begin) {};
\node[ok_node,dashed,right of=begin,xshift=-0.25cm] (goto_office) {};
\node[ok_node,right of=goto_office] (pickup1) {};
\node[ok_node,right of=pickup1] (pickup2) {};
\node[ok_node,dashed,right of=pickup2] (goto1) {};
\node[ok_node,err_node,right of=goto1] (deliver1) {};
\node[ok_node,dashed,right of=deliver1] (goto2) {};
\node[ok_node,ok_node,right of=goto2] (deliver2) {}; % Use ok_node,err_node for errors

% Nodes that are labels below states
\node[label,below of=goto_office] {$G$};
\node[label,below of=pickup1] {$P_A$};
\node[label,below of=pickup2] {$P_B$};
\node[label,below of=goto1] {$G$};
\node[label,below of=deliver1] {$D_A$};
\node[label,below of=goto2] {$G$};
\node[label,below of=deliver2] {$D_B$};

% All the arrows
\draw[ok_arr] (begin) -- (goto_office);
\draw[ok_arr] (goto_office) -- (pickup1) node[edge_label]{1};
\draw[ok_arr] (pickup1) -- (pickup2) node[edge_label]{2};
\draw[ok_arr] (pickup2) -- (goto1) node[edge_label]{3};
\draw[ok_arr] (goto1) -- (deliver1) node[edge_label]{4};
\draw[ok_arr] (deliver1) -- (goto2) node[edge_label]{8};
\draw[ok_arr] (goto2) -- (deliver2) node[edge_label]{9};
\draw[err_arr] (deliver1) edge[out=135,in=45] node[edge_label]{5} (pickup1);
\draw[err_arr] (pickup1) edge[out=40,in=150] node[edge_label]{} (goto1);
\draw[err_arr] (goto1) edge[out=30,in=150] node[edge_label]{} (deliver1);
\end{tikzpicture} \\
\scriptsize Package A missing, robot picks it up again.
\end{tabular}
\\ \hline
EL &
\renewcommand{\arraystretch}{0.2}% Tighter
\begin{tabular}{@{}l}
\rule{0pt}{3.5em}
\begin{tikzpicture}[scale=0.8, every node/.style={scale=0.8}]
% Nodes that represent states
\coordinate (begin) {};
\node[ok_node,dashed,right of=begin,xshift=-0.25cm] (goto_elevator) {};
\node[ok_node,right of=goto_elevator] (call_elevator) {};
\node[ok_node,dashed,right of=call_elevator] (enter_elevator) {};
\node[ok_node,right of=enter_elevator] (ask_floor_1) {};
\node[ok_node,dashed,right of=ask_floor_1] (wait_for_stop) {};
\node[ok_node,err_node,right of=wait_for_stop] (on_floor_1) {};
\node[ok_node,dashed,right of=on_floor_1] (exit_elevator) {};

% Nodes that are labels below states
\node[label,below of=goto_elevator] {$G$};
\node[label,below of=call_elevator] {$C_D$};
\node[label,below of=enter_elevator] {$E$};
\node[label,below of=ask_floor_1] {$A_1$};
\node[label,below of=wait_for_stop] {$W$};
\node[label,below of=on_floor_1] {$F_1$};
\node[label,below of=exit_elevator] {$E$};

% All the arrows
\draw[ok_arr] (begin) -- (goto_elevator);
\draw[ok_arr] (goto_elevator) -- (call_elevator) node[edge_label]{1};
\draw[ok_arr] (call_elevator) -- (enter_elevator) node[edge_label]{2};
\draw[ok_arr] (enter_elevator) -- (ask_floor_1) node[edge_label]{3};
\draw[ok_arr] (ask_floor_1) -- (wait_for_stop) node[edge_label]{4};
\draw[ok_arr] (wait_for_stop) -- (on_floor_1) node[edge_label]{5};
\draw[err_arr] (on_floor_1) edge[out=125,in=55] node[edge_label]{6} (ask_floor_1);
\draw[err_arr] (ask_floor_1) edge[out=20,in=160] (wait_for_stop);
\draw[err_arr] (wait_for_stop) edge[out=20,in=160] (on_floor_1);
\draw[ok_arr] (on_floor_1) -- (exit_elevator) node[edge_label]{9};
\end{tikzpicture} \\
% Person pressed the wrong floor button
\scriptsize Elevator taken to wrong floor, robot asks again.
\end{tabular}
\\ \hline
EL &
\renewcommand{\arraystretch}{0.2}% Tighter
\begin{tabular}{@{}l}
\rule{0pt}{3.5em}
\begin{tikzpicture}[scale=0.8, every node/.style={scale=0.8}]
% Nodes that represent states
\coordinate (begin) {};
\node[ok_node,dashed,right of=begin,xshift=-0.25cm] (goto_elevator) {};
\node[ok_node,right of=goto_elevator] (call_elevator) {};
\node[ok_node,err_node,dashed,right of=call_elevator] (enter_elevator) {};
\node[ok_node,right of=enter_elevator] (ask_floor_1) {};
\node[ok_node,dashed,right of=ask_floor_1] (wait_for_stop) {};
\node[ok_node,right of=wait_for_stop] (on_floor_1) {};
\node[ok_node,dashed,right of=on_floor_1] (exit_elevator) {};

% Nodes that are labels below states
\node[label,below of=goto_elevator] {$G$};
\node[label,below of=call_elevator] {$C_D$};
\node[label,below of=enter_elevator] {$E$};
\node[label,below of=ask_floor_1] {$A_1$};
\node[label,below of=wait_for_stop] {$W$};
\node[label,below of=on_floor_1] {$F_1$};
\node[label,below of=exit_elevator] {$E$};

% All the arrows
\draw[ok_arr] (begin) -- (goto_elevator);
\draw[ok_arr] (goto_elevator) -- (call_elevator) node[edge_label]{1};
\draw[ok_arr] (call_elevator) -- (enter_elevator) node[edge_label]{2};
\draw[ok_arr] (enter_elevator) -- (ask_floor_1) node[edge_label]{3};
\draw[ok_arr] (ask_floor_1) -- (wait_for_stop) node[edge_label]{7};
\draw[ok_arr] (wait_for_stop) -- (on_floor_1) node[edge_label]{8};
\draw[ok_arr] (on_floor_1) -- (exit_elevator) node[edge_label]{9};
\draw[err_arr] (enter_elevator) edge[out=125,in=55] node[edge_label]{4} (goto_elevator);
\draw[err_arr] (goto_elevator) edge[out=20,in=160] (call_elevator);
\draw[err_arr] (call_elevator) edge[out=20,in=160] (enter_elevator);
\end{tikzpicture} \\
% Person didn't call the elevator
\scriptsize Elevator not called, robot asks again.
\end{tabular}
\\ \hline
5-SC &
\renewcommand{\arraystretch}{0.2}% Tighter
\begin{tabular}{@{}l}
\rule{0pt}{3.5em}
\begin{tikzpicture}[scale=0.8, every node/.style={scale=0.8}]
% Nodes that represent states
\coordinate (begin) {};
\node[ok_node,dashed,right of=begin,xshift=-0.25cm] (goto_lab) {};
\node[ok_node,right of=goto_lab] (pickup_diss) {};
\node[ok_node,dashed,right of=pickup_diss] (goto_o1) {};
\node[ok_node,err_node,right of=goto_o1] (sign_diss) {};
\node[ok_node,dashed,right of=sign_diss] (goto_o2) {};
\node[right of=goto_o2] (elided) {...};
\node[ok_node,dashed,right of=elided] (goto_o5) {};

% Nodes that are labels below states
\node[label,below of=goto_lab] {$G$};
\node[label,below of=pickup_diss] {$P$};
\node[label,below of=goto_o1] {$G$};
\node[label,below of=sign_diss] {$S$};
\node[label,below of=goto_o2] {$G$};
\node[label,below of=goto_o5] {$G$};

% All the arrows
\draw[ok_arr] (begin) -- (goto_lab);
\draw[ok_arr] (goto_lab) -- (pickup_diss) node[edge_label]{1};
\draw[ok_arr] (pickup_diss) -- (goto_o1) node[edge_label]{2};
\draw[ok_arr] (goto_o1) -- (sign_diss) node[edge_label]{3};
\draw[err_arr] (sign_diss) edge[out=135,in=45] node[edge_label]{4} (goto_lab);
\draw[err_arr] (goto_lab) edge[out=20,in=160] (pickup_diss);
\draw[err_arr] (pickup_diss) edge[out=20,in=160] (goto_o1);
\draw[err_arr] (goto_o1) edge[out=20,in=160] (sign_diss);
\draw[ok_arr] (sign_diss) -- (goto_o2) node[edge_label]{8};
\draw[ok_arr] (goto_o2) -- (elided) node[edge_label]{9};
\draw[ok_arr] (elided) -- (goto_o5) node[edge_label]{18};
\end{tikzpicture} \\
\scriptsize Student did not give thesis; picked it up again.
\end{tabular}
\\ \hline
5-SC &
\renewcommand{\arraystretch}{0.2}% Tighter
\begin{tabular}{@{}l}
\rule{0pt}{3.5em}
\begin{tikzpicture}[scale=0.8, every node/.style={scale=0.8}]
% Nodes that represent states
\coordinate (begin) {};
\node[ok_node,dashed,right of=begin,xshift=-0.25cm] (goto_lab1) {};
\node[ok_node,right of=goto_lab1] (pickup_diss) {};
\node[unrec_node,right of=goto_lab1] (pickup_diss_x) {};
\node[ok_node,dashed,right of=pickup_diss] (goto_o1) {};
\node[ok_node,right of=goto_o1] (sign_diss1) {};
\node[right of=sign_diss1] (elided) {...};
%\node[ok_node,dashed,right of=elided] (goto_o5) {};
\node[ok_node,right of=elided] (sign_diss5) {};
\node[ok_node,dashed,right of=sign_diss5] (goto_lab2) {};
\node[ok_node,err_node,right of=goto_lab2] (deliver_diss) {};

% Nodes that are labels below states
\node[label,below of=goto_lab1] {$G$};
\node[label,below of=pickup_diss] {$P$};
\node[label,below of=goto_o1] {$G$};
\node[label,below of=sign_diss1] {$S_1$};
%\node[label,below of=goto_o5] {$G$};
\node[label,below of=sign_diss5] {$S_5$};
\node[label,below of=goto_lab2] {$G$};
\node[label,below of=deliver_diss] {$D$};

\coordinate[above of=deliver_diss,node distance=0.7cm] (anchor1);
\coordinate[above of=goto_lab1,node distance=0.7cm] (anchor2);
% All the arrows
\draw[ok_arr] (begin) -- (goto_lab1);
\draw[ok_arr] (goto_lab1) -- (pickup_diss) node[edge_label]{1};
\draw[ok_arr] (pickup_diss) -- (goto_o1) node[edge_label]{2};
\draw[ok_arr] (goto_o1) -- (sign_diss1) node[edge_label]{3};
\draw[ok_arr] (sign_diss1) -- (elided) node[edge_label]{4};
\draw[ok_arr] (elided) -- (sign_diss5) node[edge_label]{11};
\draw[ok_arr] (sign_diss5) -- (goto_lab2) node[edge_label]{12};
\draw[ok_arr] (goto_lab2) -- (deliver_diss) node[edge_label]{13};
\draw[err_arr] (deliver_diss) -- (anchor1)  -- node[edge_label]{14} (anchor2) -- (goto_lab1);
\draw[err_arr] (goto_lab1) edge[out=20,in=160] (pickup_diss);
\end{tikzpicture} \\
\scriptsize 5th committee member did not return thesis.
\end{tabular}
\\ \hline
ES &
\renewcommand{\arraystretch}{0.2}% Tighter
\begin{tabular}{@{}l}
\rule{0pt}{3.2em}
\begin{tikzpicture}[scale=0.7, every node/.style={scale=0.7}]
% Nodes that represent states
\coordinate (begin) {};
\node[ok_node,dashed,right of=begin,xshift=-0.25cm] (goto_init) {};
\node[ok_node,right of=goto_init] (ask_dest){};
\node[ok_node,right of=ask_dest] (ask_follow){};
\node[ok_node,right of=ask_follow] (escort) {};
\node[ok_node,err_node,right of=escort] (confirm_arrival) {};

% Nodes that are labels below states
\node[label,below of=goto_init] {$G$};
\node[label,below of=ask_dest] {$A$};
\node[label,below of=ask_follow] {$F$};
\node[label,below of=escort] {$E$};
\node[label,below of=confirm_arrival] {$C$};

\coordinate[above of=confirm_arrival,node distance=0.7cm] (anchor1);
\coordinate[above of=goto_init,node distance=0.7cm] (anchor2);

% All the arrows
\draw[ok_arr] (begin) -- (goto_init);
\draw[ok_arr] (goto_init) -- (ask_dest) node[edge_label]{1};
\draw[ok_arr] (ask_dest) -- (ask_follow) node[edge_label]{2};
\draw[ok_arr] (ask_follow) -- (escort) node[edge_label]{3};
\draw[ok_arr] (escort) -- (confirm_arrival) node[edge_label]{4};
\draw[err_arr] (confirm_arrival) -- (anchor1) -- (anchor2) node[edge_label]{5} -- (goto_init);
\draw[err_arr] (goto_init) edge[out=30,in=155] (ask_follow);
\draw[err_arr] (ask_follow) edge[out=20,in=160] (escort);
\draw[err_arr] (escort) edge[out=20,in=160] (confirm_arrival);
\end{tikzpicture} \\
\scriptsize Visitor did not arrive at destination; escort restarted.
\end{tabular}
\\ \hline
ES &
\renewcommand{\arraystretch}{0.2}% Tighter
\begin{tabular}{@{}l}
\rule{0pt}{3.5em}
\begin{tikzpicture}[scale=0.8, every node/.style={scale=0.8}]
% Nodes that represent states
\coordinate (begin) {};
\node[ok_node,dashed,right of=begin,xshift=-0.25cm] (goto_init) {};
\node[ok_node,right of=goto_init] (ask_dest){};
\node[ok_node,right of=ask_dest] (ask_follow){};
\node[unrec_node,right of=ask_dest] (ask_follow_x){};
\node[ok_node,right of=ask_follow] (escort){};
\node[ok_node,err_node,right of=escort] (confirm_arrival){};

% Nodes that are labels below states
\node[label,below of=goto_init] {$G$};
\node[label,below of=ask_dest] {$A$};
\node[label,below of=ask_follow] {$F$};
%\node[edge_label,below of=ask_follow_x] {7};
\node[label,below of=escort] {$E$};
\node[label,below of=confirm_arrival] {$C$};

\coordinate[above of=confirm_arrival,node distance=0.7cm] (anchor1);
\coordinate[above of=goto_init,node distance=0.7cm] (anchor2);

% All the arrows
\draw[ok_arr] (begin) -- (goto_init);
\draw[ok_arr] (goto_init) -- (ask_dest) node[edge_label]{1};
\draw[ok_arr] (ask_dest) -- (ask_follow) node[edge_label]{2};
\draw[ok_arr] (ask_follow) -- (escort) node[edge_label]{3};
\draw[ok_arr] (escort) -- (confirm_arrival) node[edge_label]{4};
\draw[err_arr] (confirm_arrival) -- (anchor1) -- (anchor2) node[edge_label]{5} -- (goto_init);
\draw[err_arr] (goto_init) edge[out=30,in=155] (ask_follow);
\end{tikzpicture} \\
\scriptsize Visitor did not arrive at destination; lost.
\end{tabular}
\\ \hline
\end{tabular}
\caption{Execution traces of several task programs. Nodes are actions and edges
represent control flow. The filled, red nodes are actions that failed, and the
red edges depict control flow during error recovery. Crossed out nodes represent
unrecoverable failures.}
\tablabel{traces}
\end{table}

\subsection{Variation in Failure Recovery}
\seclabel{eval1}

 To evaluate the ability of \rtpl{} task programs to recovery from a variety
 of failures, we wrote four task programs for a service mobile robot using
 \rtpl{}:
 \begin{enumerate}

\item n-package delivery (\textbf{n-PD}) : the robot picks up $n$ packages from
the mail room and delivers them to $n$ recipients. 
The human interactions are to 1)~pickup a package from
the mailroom and 2)~give a package to its recipient.

\item Elevator (\textbf{EL}): the robot takes the elevator from one floor to
another with human assistance. The human interactions are to
1)~call the elevator, 2)~press a button to take the robot to a floor,
3)~hold the elevator door open for the robot, and 4)~confirm that the robot is
on the right floor.

\item n-signature collection (\textbf{n-SC}) : the robot picks up the
manuscript for a thesis from a student's office, and collects the signatures
from $n$ thesis committee members. The human interactions are
to 1)~pickup the thesis from the student, 2)~give the thesis to a committee
member to sign, and 3)~take back the thesis from a committee member.

\item Escort (\textbf{ES}): the robot escorts a visitor between locations. The
human interactions are to ask the visitor 1)~for their destination,
2)~to start following the robot, 3)~stay with the robot, and 
4)~confirm that they have arrived.

\end{enumerate}
These programs do not have explicit failure-recovery logic
present.\footnote{The full code listings for all programs are in Appendix~\ref{apx:task-programs}
the supplemental material.} Therefore, without \rtpl{} every
programs would go wrong if any action or human interaction failed.

For this experiment, we execute each task program multiple times with different
failures, thus witness different kinds of automatic failure recovery.
\tabref{traces}. For every combination of program and failure, we graphically
depict the execution trace of the task. In each graphic, the nodes represent both
autonomous actions (dashed border) and actions where the robot asks humans
for help  (solid border). The filled, red nodes are
actions where a failure is detected. The black edges indicate normal program
execution, whereas the red edges indicate the flow of control during failure
recovery. For clarity, we number several edges to better depict the sequence
in which actions are performed.

These experiments show a variety of different of failure recovery behaviors
that \rtpl{} permits for several task programs, which \emph{all use the same
robot model}. We show different failures for the same program, because they
illustrate the difficulty of accounting for all possible errors, particularly
for non-experts. Therefore, \rtpl{} thus spares non-experts from having to
reason through all possible failures themselves. Note that the execution traces
involve jumping back several steps to retry and action and jumping over actions
that succeeded. To write a program that can jump back and forth in this manner
requires significant programming expertise: the ability to write several nested
loops with complex exit conditions or many auxiliary functions.

\begin{figure}[t]
  \centering
  \hspace{-1.5cm}
  \begin{subfigure}[b]{0.49\linewidth}
     \centering
     \includegraphics[width=\columnwidth]{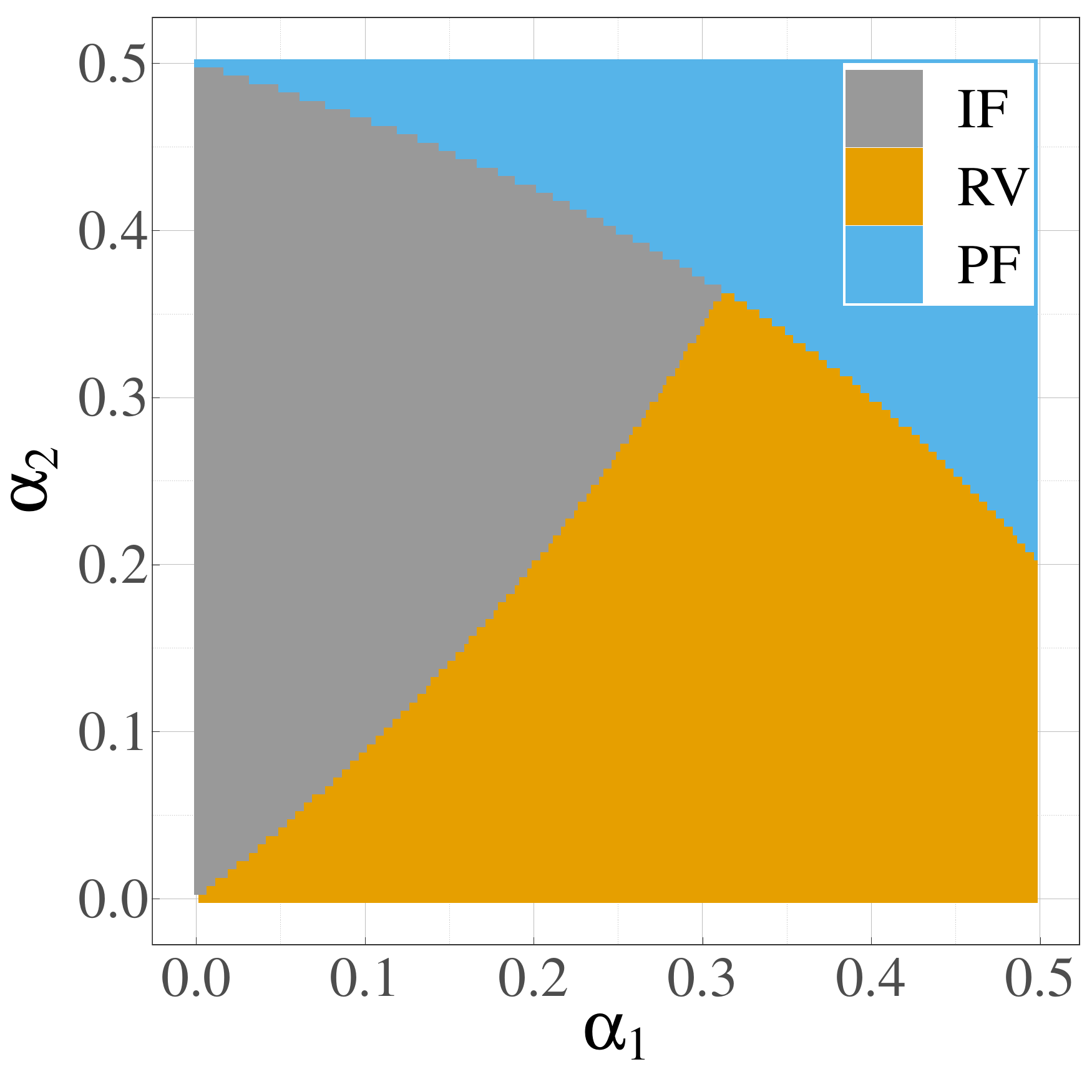}
     \caption{ES Program.}
     \figlabel{es-phase}
  \end{subfigure}
  \hspace{-0.125cm}
  \begin{subfigure}[b]{0.49\linewidth}
     \centering
     \includegraphics[width=\columnwidth]{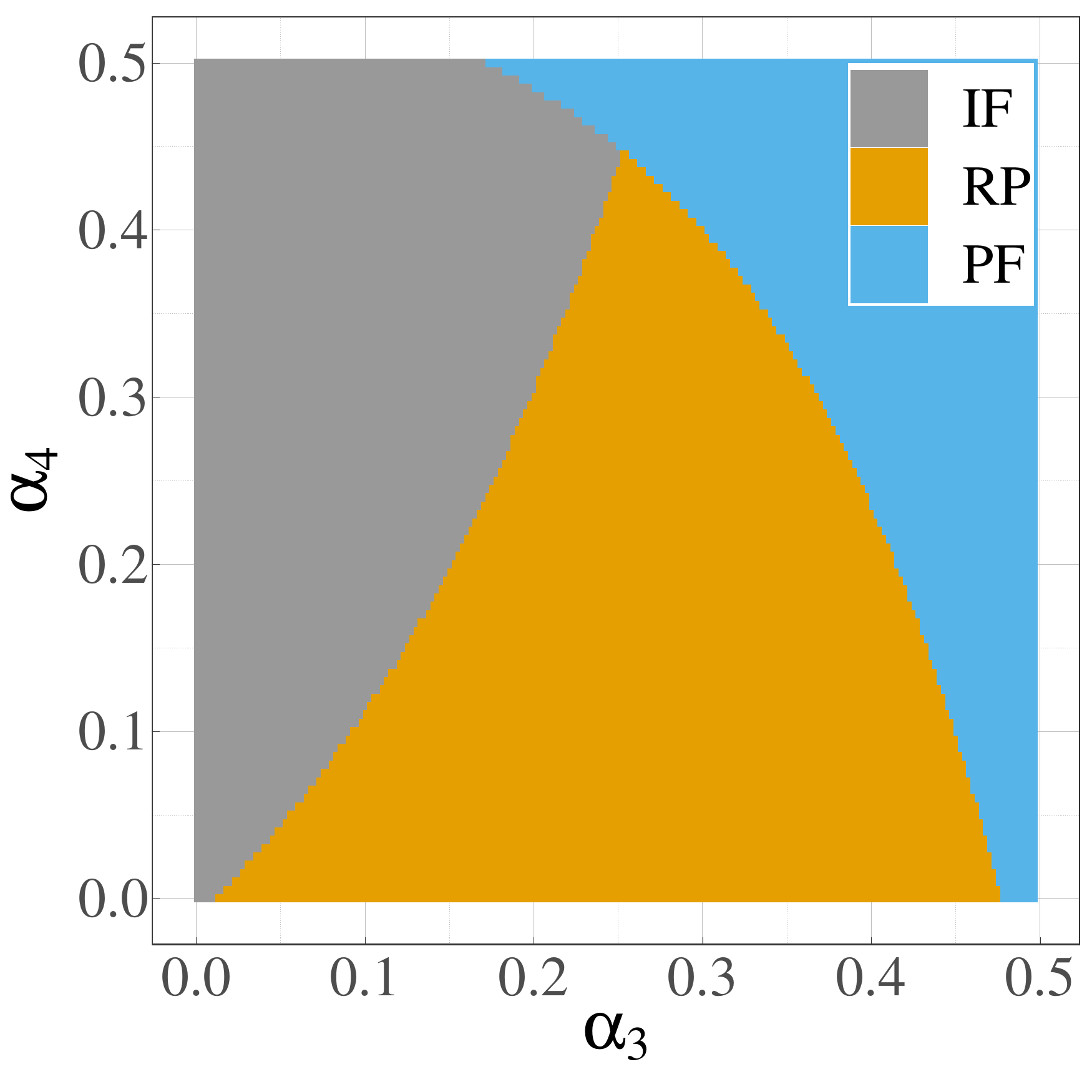}
     \caption{2-PD Program.}
     \figlabel{pd-phase}
   \end{subfigure}
   \hspace{-1.5cm}
   \vspace{-.7em}
   \caption{Effect of varying robot model parameters on recovery behavior in different programs.}
   \figlabel{phase}
 \end{figure}

\subsection{The Effect of Model Parameters}
\seclabel{eval2}

An \rtpl{} robot-model uses parameters that determine the
prior probability of various outcomes, including failures. The
values of these parameters affect the behavior of task programs during failure
recovery and cause \rtpl{} to produce different execution traces. Therefore, it
is important for the roboticist building the robot model to understand how
these parameter values affect failure recovery.
We investigate the impact of parameter values on the visitor escort program (ES) and the two-package delivery
program (2-PD) in a simulated experiment with the following human interactions as
ground truth: 
\begin{inparaenum}[1)]
\item In ES, the visitor confirms that they are going to
follow the robot, but does not confirm arrival at the destination, which
triggers failure recovery.
\item In t2-PD, the mailroom tells the robot that both
packages have been given, but the human at destination B tells the
robot that package B is missing, which triggers failure recovery.
\end{inparaenum}

\figref{phase} plots the possible outcomes where
vary the parameters
\begin{inparaenum}[1)]
\item $\alpha_1$, the probability that the person fails to follow the robot after
 the \lstinline|askFollow| action;
\item $\alpha_2$, the probability that the person loses track of the robot during the \lstinline|escortTo| action;
\item $\alpha_3$, the probability that the human fails to give the robot the
right item during the package pickup action \lstinline|pickup|; and
\item $\alpha_4$, the probability that a human takes a wrong package from the
robot during the package delivery action \lstinline|give|.
\end{inparaenum}
Note that outcomes for parameter values $>0.5$ are not plotted, since they imply
that the action is more likely to fail than not, which would be autonomously
caught as fatal unrecoverable errors (\secref{failure-recovery}).

\figref{es-phase} shows the three possible outcomes of the ES program:
\begin{inparaenum}[1)]
  \item for small values of $\alpha_2$ compared to $\alpha_1$, the robot infers that
  the most likely cause of failure was that the visitor was left behind at the
  start, thus the robot going back to the start to
  re-engage the visitor (RV). 
\item For small values of $\alpha_1$ and larger values of $\alpha_2$, the robot
expects that the visitor is still with the robot at the destination. When the
confirmation action times out, it results in an inferred unrecoverable failure (IF) when
it infers that the visitor most likely stopped
following the robot en route. 
\item For larger values of $\alpha_1$ and $\alpha_2$, the robot
 encounters a predicted failure (PF) when it predicts at the destination
 that the person is most likely no longer with the robot.
\end{inparaenum}

\figref{pd-phase} shows the three possible outcomes of the 2-PD program:
\begin{inparaenum}[1)]
  \item For small values of $\alpha_4$ compared to $\alpha_3$, the robot infers that
the most likely cause of failure was that the human did not actually give the
package to the robot during pickup, hence the robot
goes back to the mail room  to re-pickup the package (RP). 
\item For small values of $\alpha_3$ and larger values of $\alpha_4$, the robot
expects that package B is still with the robot at destination B. However, when
person B indicates that the robot does not have the package, it
 results in an inferred unrecoverable failure (IF) when it infers that the package
 was lost in transit.
\item For larger values of both $\alpha_3$ and $\alpha_4$, the robot
 encounters an unrecoverable predicted failure (PF) when it predicts at
destination B that package B was most likely taken from the robot in transit.
\end{inparaenum}

\begin{table}
\centering
\footnotesize
\begin{tabular}{|l|l|r|r|}
\hline
\thead{Task} & \thead{Failure} & \thead{\rtpl{}} & \thead{Re-Execution} \\
\hline
% (not (have ?package-2))
2-PD & Package 2 delivery fails & 4m33s & 7m01s \\
% (not (have ?package-3))
3-PD & Package 2 delivery fails & 6m31s & 7m17s \\
% (not (selected ?floor-1))
EL & Wrong floor selected & 1m31s & 2m28s \\
\hline
\end{tabular}
\caption{Failure recovery time for \rtpl{} vs. re-execution.}
\tablabel{real-robots}
\end{table}

\subsection{Real-World Execution Time}
\seclabel{eval3}

A robot task program that neither uses \rtpl{} nor has explicit failure
recovery code can be re-executed in full when a failure occurs.
For certain failures, a complete
re-execution may be undesirable. For example, if a program to deliver two
packages fails to deliver one of them, then a naive re-execution where it
attempts to deliver an already-delivered package will annoy humans. Therefore,
failure recovery, whether implicit with \rtpl{} or explicit, is necessary for
service mobile robots to behave in socially acceptable ways. Setting concerns
about social acceptability aside, naive re-execution is a baseline
to measure how much time \rtpl{} saves by only re-executing a subsequence
of actions.
\tabref{real-robots} shows the results of these experiments on three programs
to deliver two packages (\textbf{2-PD}), deliver three packages
(\textbf{3-PD}), and use the elevator (\textbf{EL}). The table describes the
kind of failure induced in each experiment, along with the time taken with
\rtpl{} and naive re-execution. In all cases, using \rtpl{} is faster than
a full re-execution.

% =============================================================================
\section{Conclusion}

This paper presents \rtplfull{} (\rtpl{}), a language for programming novel tasks for service
mobile robots. \rtpl{} allows end-user programs to be written in a simple
sequential manner, while providing autonomous failure inference and recovery.
We demonstrate that \rtpl{}:
\begin{inparaenum}[1)]
\item~allows complex tasks to be written concisely, 
\item~correctly identifies the root cause of failure, and 
\item~allows multiple tasks to recover from a variety of errors, without task-specific error-recovery code.
\end{inparaenum}

\bibliography{main}
\bibliographystyle{aaai}

% =============================================================================
% =============================================================================
\newpage

\appendix

% =============================================================================
\section{Actions}

The following table lists the interactions we have built with \rtpl{}
and describes their nominal specifications and error models.

\begin{table}[h]
% NOTE(arjun): Do not use \begin{table}.
\footnotesize
\lstset{basicstyle=\fontsize{8}{6.5}\ttfamily}
\begin{tabular}{@{}p{0.35\columnwidth}|@{\,}p{0.275\columnwidth}|@{\,}p{0.3\columnwidth}}
\hline
\textbf{Name} & \textbf{Nominal} & \textbf{Error Model} \\
              & \textbf{Behavior} &   \\
\hline
\lstinline|pickup(X)|
& Robot asks human to place $X$ its basket
& Robot does not receive $X$, or receives some other item \\
\hline
\lstinline|give(X)|
& Robot asks human to take $X$ from its basket
& Human does not take $X$, or takes the wrong item \\
\hline
\lstinline|getSignature(X)|
& Robot asks human to sign $X$ and return it to its basket
& Human does not sign $X$, or does not return $X$ \\
\hline
\lstinline|callElevator(X)|
& Robot asks human to call the elevator using X button
& Human does not press X to call the elevator \\
\hline
\lstinline|selectFloor(X)|
& Robot asks human to select floor $X$ inside the elevator
& Human does not press the button for floor $X$ \\
\hline
\lstinline|confirmFloor(X)|
& Robot asks human if the elevator is at floor $X$
& Human incorrectly reports that the elevator is on floor $X$ \\
\hline
\lstinline|askFollow(X)|
& Robot asks human to accompany it 
& Human does not follow the robot, or fails to respond promptly \\
\hline
\lstinline|escortTo(X)|
& Robot goes to $X$ accompanied by a human 
& Human does not stay with the robot to their destination $X$ \\
\hline
\lstinline|confirmArrival(X)|
& Robot asks human to confirm their arrival at destination $X$
& Robot is not at $X$, or human fails to respond promptly \\
\hline
\end{tabular}
\caption{Summary of human-robot interactions that we have built in \rtpl{}.}
\end{table}

% =============================================================================
\newpage
\section{Robot Task Programs}
\label{apx:task-programs}

In the elevator program (EL), the service mobile robot takes the elevator to
the first floor, with human assistance:

\begin{figure}[h]
\lstset{language=python}
\hrule
\begin{lstlisting}
robot.goto("elevator")
robot.callElevator("down")
robot.enterElevator()
robot.selectFloor(1)
robot.waitForElevatorStop()
robot.confirmFloor(1)
robot.exitElevator(1)
\end{lstlisting}
\hrule
\end{figure}

\vspace{2em}
\noindent
In the escort visitor (ES) program, the robot escorts a visitor to one of three rooms
in a building:
  
\begin{figure}[h]
\hrule
\lstset{language=python}
\begin{lstlisting}
robot.goto("initial location")
destination = robot.prompt("Which room are you looking for?", buttons = ["A323", "A325", "A327"])
robot.askFollow("initial location")
robot.escortTo(destination)
robot.confirmArrival(destination)
\end{lstlisting}
\hrule
\end{figure}

\vspace{2em}
\noindent
In the n-package delivery (n-PD) program, the robot picks up $n$ packages from a mail room
and delivers them to their recipients:

\begin{figure}[h]
\hrule
\lstset{language=python}
\begin{lstlisting}
robot.goto("mail room")
for num in range(n):
  item_name = f"package {num}"
  robot.pickup(item_name)
for num in range(n):
  delivery_loc = f"office {num}"
  item_name = f"package {num}"
  robot.goto(delivery_loc)
  robot.give(item_name)
\end{lstlisting}
\hrule
\end{figure}

\vspace{2em}
\noindent
In the n-signature collection (n-SC) program, the robot takes a thesis to $n$ committee members,
asks each one to sign it, and then returns the thesis to the student:

\begin{figure}[!h]
\hrule
\lstset{language=python}
\begin{lstlisting}
def collectSignature(num):
  sig_name = f"signature {num}"
  office = offices[num]
  robot.goto(office)
  robot.getSignature(office, sig_name, "dissertation")

robot.goto("lab")
robot.pickup("dissertation")
for num in range(n):
  collectSignature(num)
robot.goto("lab")
robot.give("dissertation")
\end{lstlisting}
\hrule
\end{figure}

\end{document}